\documentclass[letterpaper, 10 pt, journal, twoside]{IEEEtran}
\usepackage{graphicx}
\usepackage{amsmath} 
\usepackage{threeparttable}
\usepackage{array}
\usepackage{float}
\usepackage{subfigure}
\usepackage{cite}
\usepackage[numbers,sort&compress]{natbib}
\usepackage{array} 
\usepackage{multirow}
\usepackage{amsfonts}
\usepackage{booktabs} 
\usepackage{amsthm} 

\usepackage{hyperref} 
\hypersetup{
	colorlinks=true,
	linkcolor=blue,
	filecolor=blue,      
	urlcolor=blue,
	citecolor=blue,
}
\usepackage{bm}
\usepackage{color}
\usepackage[table]{xcolor}

\ifCLASSINFOpdf
\else
\fi
\begin{document}
%
\title{Image-to-Force Estimation for Soft Tissue Interaction in Robotic-Assisted Surgery Using Structured Light}
%
%
%


\author{Jiayin Wang$^{1,2}$, Mingfeng Yao$^{2}$, Yanran Wei$^{*3}$, Xiaoyu Guo$^{4}$, Ayong Zheng$^{2}$ and Weidong Zhao$^{*1}$%



\thanks{$^{1}$School of Electronic Information Engineering, Tongji University, 200092, Shanghai, China.}%
\thanks{$^{2}$MicroPort MedBot (Group) Company Ltd., 201203, Shanghai, China.}%
\thanks{$^{3}$College of Engineering, Peking University, 100871, Beijing, China.}%
\thanks{$^{4}$Department of Biomedical Engineering, City University of Hong Kong, 999077, Kowloon, Hong Kong.}%
\thanks{\textsuperscript{*} Corresponding author.}
\thanks{E-mail:\{wangjy,MingFeng.Yao,ayzheng\}@microport.com,yrweibuaa@126.com, xiaoyguo@cityu.edu.hk, wd@tongji.edu.cn}

}
\maketitle

\begin{abstract}
For Minimally Invasive Surgical (MIS) robots, accurate haptic interaction force feedback is essential for ensuring the safety of interacting with soft tissue. However, most existing MIS robotic systems cannot facilitate direct measurement of the interaction force with hardware sensors due to space limitations. This letter introduces an effective vision-based scheme that utilizes a One-Shot structured light projection with a designed pattern on soft tissue coupled with haptic information processing through a trained image-to-force neural network. The images captured from the endoscopic stereo camera are analyzed to reconstruct high-resolution 3D point clouds for soft tissue deformation. Based on this, a modified PointNet-based force estimation method is proposed, which excels in representing the complex mechanical properties of soft tissue. Numerical force interaction experiments are conducted on three silicon materials with different stiffness. The results validate the effectiveness of the proposed scheme.
\end{abstract}


%
\IEEEpeerreviewmaketitle

\section{Introduction}

\IEEEPARstart{M}{inimally} Invasive Surgery (MIS) robotic systems represent a formidable frontier in contemporary medicine, offering reduced tissue trauma and improved operational safety~\cite{abdi2020haptics}. However, these systems often prohibit direct haptic sensing between the surgeon and soft tissues, thereby increasing the risk associated with real-time force interactions. Therefore, haptic force sensing in such scenarios has become an essential requirement~\cite{juo2020center,cabibihan2021influence}.

The primary methods for haptic force sensing include~\cite{puangmali2008state}: additive force sensor-based measurement and sensorless force estimation. In~\cite{hadi2022force,kim2016development,rosen1999force}, force sensors are mounted on the end-effectors of surgical robots to directly measure interaction forces. Alternatively, in~\cite{bahar2020surgeon}, sensors are affixed to the surface of the tissue itself. While these methods provide intuitive operation and high measurement accuracy, their clinical application remains hindered by challenges such as cost constraints, limited installation space, and inadequate resistance to high temperatures and corrosion~\cite{calandra2018more}.

To address these limitations, sensorless force estimation methods have been developed. In previous work~\cite{TASE2023_Wei_force}, the dynamic information of the robot was utilized to estimate external interaction forces. In~\cite{TM2024_Wei}, the mechanical properties of the deformable environment were integrated with the robot dynamics to improve the accuracy of the estimation. Although these dynamic models-based robotic methods are both effective and non-reliant on additive sensors, they inherently rely on precise modeling of the dynamics of the surgical robot. Another indirect force estimation approach is vision-based force estimation (VBFE) methods which refers the force from the model of deformable objects and the displacement of the surface. In~\cite{obinata2007vision}, a method to predict surface force and friction coefficients by embedding marked elastomers in silicone membranes was proposed. However, model-based VBFE methods are not appropriate for real-time applications due to the requirement of inaccessible a priori knowledge of the reference shape and the mechanics information~\cite{Sensor_J2021_VisionReview,vlack2005gelforce}. 

Utilizing the versatility of deep learning methods to model complex deformation, learning-based VBFE methods are developed~\cite{mirniazy2022supervised,takahashi2019deep,pecyna2022visual}. In~\cite{naeini2019novel}, a force estimation method is proposed via time-delayed neural networks and Gaussian processes based on dynamic vision sensors. In~\cite{aviles2015sensorless}, the surface deformation is modeled using cubic B-splines combined with an energy minimization strategy, while the visual-geometric-force relationship is learned through a recurring neural network (RNN). However, a significant limitation of this approach lies in the absence of a detailed dataset for training, which is challenging to obtain in medical applications. Moreover, the aforementioned methods are primarily limited to push actions, overlooking the more complex force estimation required for pull (traction) tasks, which represent a particularly challenging scenario in MIS systems.

Towards the goal of practical vision-based force estimation in MIS systems, two major technical challenges arise: 1) How to establish a vision-based force estimation framework suitable for scenarios where a surgical robot interacts with texture-deficient soft tissues, particularly during pulling tasks, which are both more challenging and common in surgical procedures (e.g., suturing, cutting); 2) How to model the complex displacement-force relationship and train it using a high-quality custom dataset specifically designed for this task, acknowledging the well-known difficulty in collecting datasets for physical interaction in medical contexts.

This letter proposes a novel VBFE scheme that leverages structured light projection to actively characterize the 3D surface of texture-deficient soft tissue and constructs point-cloud models using stereo vision, a common configuration in medical endoscopes. A modified PointNet-based method is developed to learn the displacement-force relationship offline, utilizing a custom dataset specifically designed for this task. The effectiveness of the proposed method has been validated on the commercial Toumai laparoscopic surgical robot platform. The main contributions are summarized as follows:
 
\begin{itemize}
\item [(1)] A novel VBFE framework designed specifically for laparoscopic MIS robots is proposed. This framework employs point clouds for 3D representation and enhances the existing PointNet-based network to learn the displacement-force relationship in an offline manner. Due to the smooth and texture-deficient surface of human tissues, it is challenging to directly utilize point clouds to represent the deformation of soft tissue in MIS applications, as an insufficient number of points will lead to failed stereo vision matching. To this end, an active approach is implemented that uses structured light projection with fringe patterns to enhance surface texture. In this way, dense (pixel-wise) point clouds can be obtained, allowing for high-resolution 3D reconstruction. In contrast to traditional force estimation methods, the framework eliminates the need for additional sensors, depth cameras, and prior knowledge of the mechanical properties of the materials. This enhances both accuracy and generalizability while leveraging the inherent data from the endoscope for MIS platforms. 

\item [(2)]A modified PointNet-based force estimation method is proposed to enhance the process of characterizing the displacement force model from the dataset. This deep learning-based force estimator has been improved in three key areas compared to the original PointNet: input data preprocessing, optimization of the displacement-force model, and refinement of the loss function for training. Unlike traditional methods, this approach incorporates additional input features using active 3D reconstruction based on structured light projection and point clouds generated through stereo vision matching. This enhancement increases the robustness of feature recognition in the presence of variations in illumination, noise, and image distortion. Furthermore, this method retains the original PointNet network architecture but uses the exponential linear unit (ELU) as the activation function. The output layer is modified to suit regression tasks, and the Nesterov-accelerated adaptive moment (Nadam) estimation algorithm is utilized as the optimizer for rapid training. The experimental results validate the accuracy and effectiveness of the proposed scheme.
\end{itemize}

The outline of the letter is as follows. Section II outlines the necessary preliminaries and defines the problem. Section III details the proposed VBFE framework. Section IV presents experimental results from a real-world force interaction task conducted with the Toumai laparoscopic MIS robots. Finally, Section V concludes this letter. 

\section{Preliminaries} \label{II}
\subsection{Force Model for Deformable Tissue}
Classical constitutive models are used to describe the deformation behavior of different materials under external interaction forces, as follows
\begin{equation}\label{forcemodel}
\sigma=f(\epsilon),
\end{equation}
\noindent where $\sigma$ and $\epsilon$ represent the stress and strain of the deformable object, respectively. Constitutive models based on the mechanical assumptions of the materials are used to infer the interaction force estimate including the elastic models, hyperelastic models, and viscoelastic models.

A kind of accurate model for capturing the time-varying behavior of the soft tissues is viscoelastic models including the Maxwell model and the Kelvin-Voigt model. The two models can effectively describe the stress relaxation behavior and the creep behavior of viscoelastic materials, respectively. These two models are formulated as Eq.~(\ref{Max}) and Eq.~(\ref{KV}).

\begin{equation}\label{Max}
\frac{d \epsilon}{d t} = \frac{1}{E} \times \frac{d \sigma}{d t} + \frac{\sigma}{\eta},
\end{equation}

\begin{equation}\label{KV}
\sigma \left(t\right) = E \epsilon + \eta \frac{d \epsilon}{d t},
\end{equation}
\noindent where $\eta$ and $E$ are the viscosity coefficient and elastic modulus, respectively. In surgical simulations, constitutive models are formulated using Finite Element Analysis (FEA) tools to decompose displacement fields and analyze soft tissue deformations accurately. However, FEA is known to be complex and computationally expensive, which limits the application of online VBFE in surgical robotics. In addition, the accuracy of the model-based force estimation methods is highly based on prior knowledge of mechanical parameters that cannot be easily obtained. 

\subsection{Vision Reconstruction}
VBFE methods in this letter are based on image-based vision reconstruction. Dense point clouds are generated using active or passive sensing techniques to facilitate high-resolution 3D reconstruction, providing detailed descriptions of deformable object shapes. In most laparoscopic surgery scenarios, endoscopes are equipped with ordinary stereo cameras, such as the commercial endoscope (DFVision) highlighted in this letter. These cameras do not require active light sources; instead, they capture images of the same scene from different viewpoints (i.e., left and right views) using two cameras. Depth information is computed by matching feature points in the images and applying triangulation methods. This approach is well-suited for scenes with abundant details; however, in low-texture or textureless regions (e.g., smooth soft tissue surfaces), depth computation may fail due to the difficulty in finding matching points.

Coded structured light-based shape reconstruction is a reliable active technique to recover the surfaces of objects. This approach effectively generates artificial features on smooth surfaces. Structured light systems can achieve high accuracy and, with appropriate algorithmic optimization, enable real-time depth estimation, particularly in short-range and low-speed motion scenarios. 


\subsection{Problem Formulation}
In laparoscopic surgery, the task of applying a unidirectional, low-speed pulling force is challenging and should be prioritized to minimize the risk of unexpected soft tissue damage~\cite{shirk2006complications}. For simplicity, and without loss of generalization, this scenario makes two fundamental assumptions about the deformable object: 1) its mechanical properties are isotropic, and 2) its geometric properties are uniform, disregarding minor geometric irregularities.

Classical methods of VBFE for deformable objects using binocular stereo vision typically follow this workflow: First, a disparity map is generated via stereo matching. This map is then used to reconstruct 3D point clouds representing the object's surface. Finally, a constitutive displacement-force model is derived using Finite Element Analysis (FEA) and the material's mechanical properties. However, these methods fail in this scenario due to: 1) The nonlinearity of soft tissue mechanics, which complicates accurate modeling; 2) The texture deficiency of soft tissue surfaces, impairing the accuracy of visual reconstruction and leading to force estimation errors; 3) Poor real-time performance, which fails to meet the demands of surgical applications. 

To achieve accurate and computationally fast force estimation, One-Shot-based pattern projection is required to achieve dense (pixel-wise) reconstruction, absolute coding, and high accuracy; in addition, a deep learning-based displacement force model, as an alternative to mechanistic modeling in Eq.~(\ref{forcemodel}), is established by directly processing point cloud data without relying on structured representations such as voxels or meshes. 


\section{The Presented Scheme}
In this section, the proposed VBFE scheme is detailed. In Section III-A, we present a stereo vision 3D reconstruction method for deformable tissue in MIS using a designed One-Shot absolute structured light projection. In Section III-B, a modified PoinNet-based force estimation network is presented. 
\subsection{3D Reconstruction with One-Shot Structured Light} 
The 3D reconstruction process in this scheme involves designing a specialized One-Shot absolute structured light pattern, performing stereo vision matching using the SGBM algorithm with pattern projection, and generating a real-time dense 3D point cloud of the object surface.

\subsubsection{Structured Light Pattern Creation} 

\begin{figure}[tp]
	\centering
\includegraphics[width=0.9\linewidth]{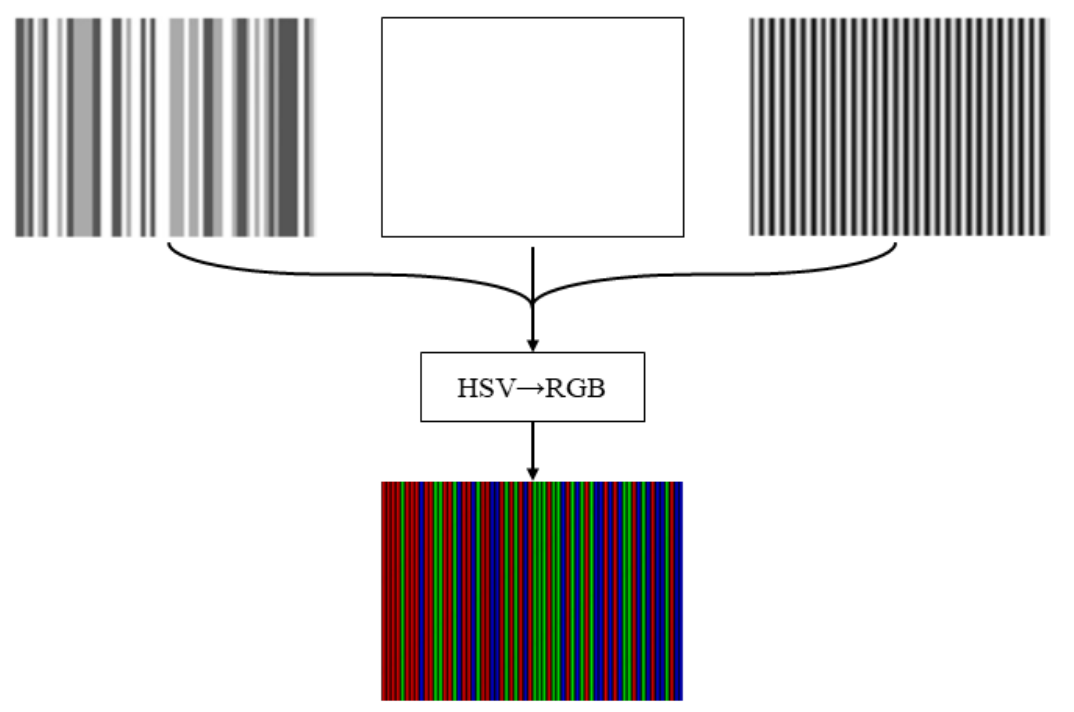}
\caption{Pattern creation. H channel pattern generated by De Bruijn sequence (Top-left); S channel with constant maxima (Top-middle); Sinusoidal intensity pattern in V channel (Top-right); The result RGB pattern.}
\label{project}
\end{figure}

\begin{figure}[tp]
	\centering
	\subfigure[]
	{
		\centering
		\includegraphics[width=0.4\linewidth]{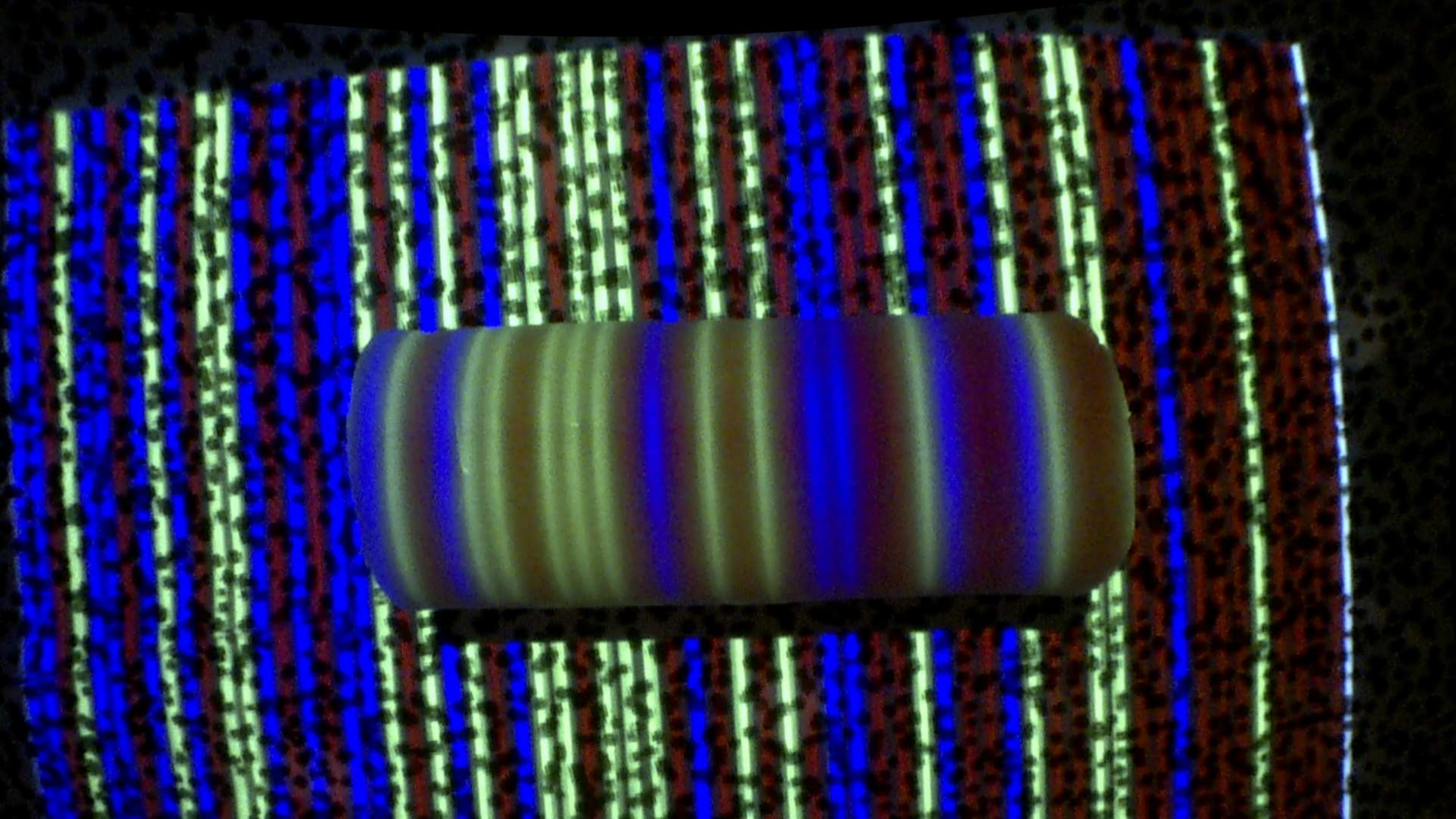}
	}
	\subfigure[]
	{
		\centering
		\includegraphics[width=0.4\linewidth]{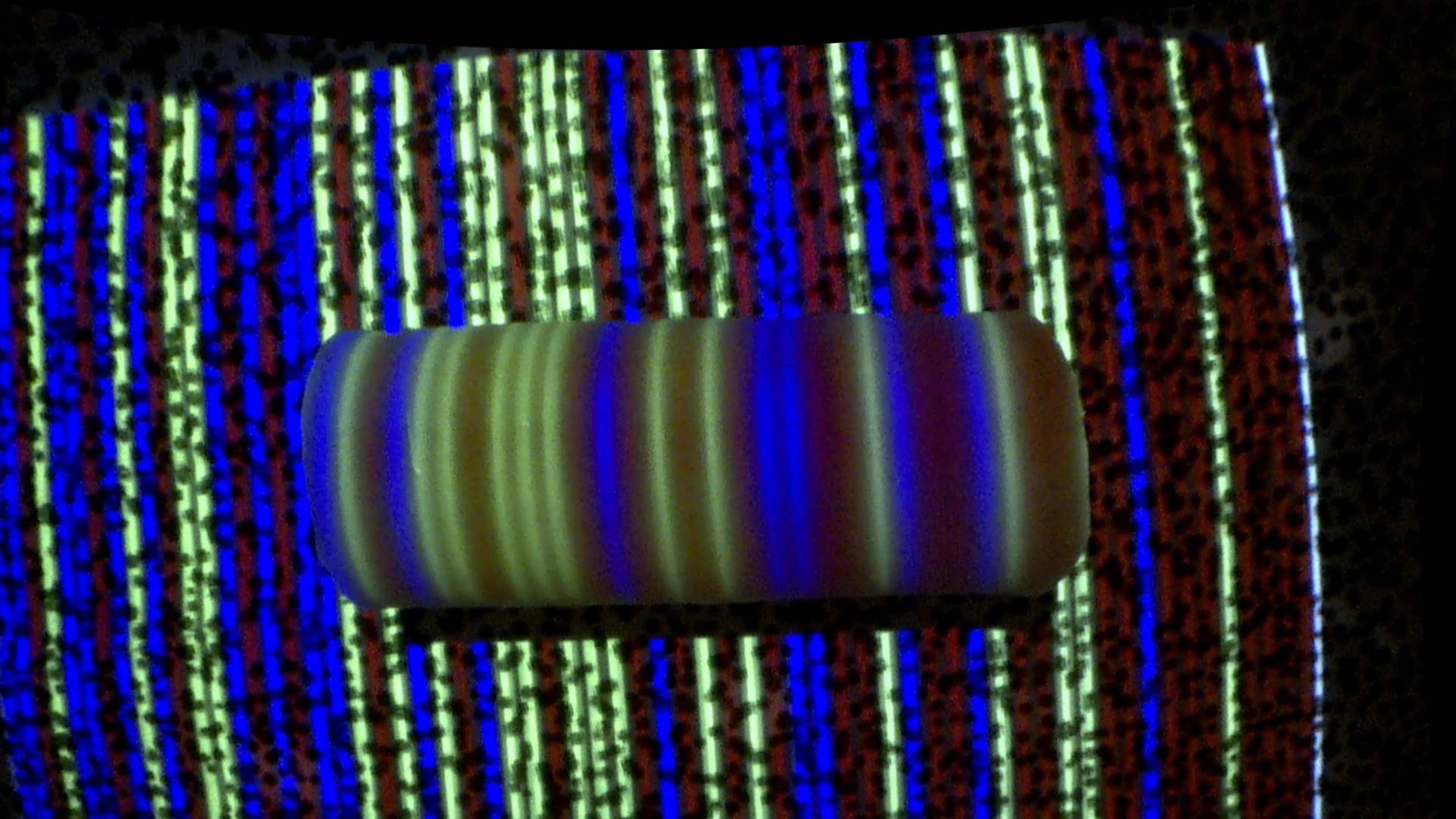}
	}
	\caption{Images of structured light projection captured by the stereo camera system: (a) left camera image, (b) right camera image.}\label{touying}
  \end{figure}

To achieve time-efficient reconstruction, a One-Shot absolute pattern is employed for structured light encoding. As shown in Fig.~\ref{project}, the pattern consists of a set of colored sinusoidal fringes generated in the Hue, Saturation, and Value (HSV) color space, with the H channel encoded using the DeBruijn sequence. A DeBruijn sequence is a circular sequence in which each element belongs to an alphabet of \( n \) symbols. This sequence can be directly constructed from the Hamiltonian or Eulerian paths of an \( n \)-dimensional DeBruijn graph~\cite{fredricksen1982survey}. A key property of the DeBruijn sequence is that any substring of length \( m \) appears exactly once, making it ideal for generating the colored fringe sequence in the H channel. This unique encoding property ensures that each structured light unit corresponds uniquely to its decoded information, thereby enhancing the accuracy of stereo matching. 

In the proposed scheme, the alphabet comprises \(n=3\) symbols: red, green, and blue. Based on the projector resolution, a fringe sequence of length 64 is selected as the encoding pattern. The substring length \( m \) is correspondingly set to 4, generating a DeBruijn sequence of length 81 to satisfy the condition \(n^{m}>64\). The saturation of the pattern is set to its maximum for all pixels, which means that the S channel is fixed at 1 for every pixel. The vertical (top-to-bottom) light intensity of each colored fringe follows a sinusoidal distribution. Consequently, in the V channel, the light intensity signal for each column is represented as:
\begin{equation}
\begin{split}
I\left(i\right) = 0.5 + 0.5 \cdot \cos\left(2\pi f \cdot i\right),\ i=1\ldots N,
\end{split}
\end{equation}
\noindent where \( i \) denotes the column index, \( N = 64 \) represents the maximum horizontal resolution of the projector, and \( f \) is the frequency given by \( f = \frac{64}{N} \). 

The structured light pattern generated in the HSV color space is transformed into the RGB space and projected onto the silicone model of a humanoid intestine. The images captured by the stereo camera are shown in Fig.~\ref{touying}, and structured light is then used for stereo matching.
\subsubsection{Pattern Recovery and Stereo Matching} 
Before decoding, the base color of the object is removed and the pixel correction is performed using the Caspi model. To enhance stereo-matching accuracy, this letter adopts a two-step approach comprising initial matching and subsequently refined matching verification between matched point pairs. 

For initial matching, the SGBM algorithm~\cite{hirschmuller2007stereo} is used to generate candidate-matching pairs and calculate their encoded information. In the refined validation step, the DeBruijn decoded information of each pair of points of interest (POI) is verified. The pair is considered mismatched if the decoded DeBruijn subsequences differ between the left- and right-camera views. A search is then conducted within the candidate SGBM match set and the neighborhood of the original match points to locate a matching point with the same value.



First, the horizontal Sobel operator is applied to preprocess the left and right images. Specifically, for each pixel $p$ in the paired images, the matching cost is calculated as follows:
\begin{equation}
C \left(p , d\right) = C_{original}^{BT} \left(p , d\right) + C_{sobel}^{BT} \left(p , d\right),
\end{equation}
\noindent where $C_{original}^{BT} \left(p, d\right)$ represents the Birchfield-Tomasi cost of pixel $p$ in the original image, and $C_{sobel}^{BT} \left(p, d\right)$ represents the Birchfield-Tomasi cost of pixel after pre-processing. \( d \) is the disparity value. The features are stored in a disparity space image (DSI) matrix for subsequent matching.

\begin{figure}[tp]
	\centering
\includegraphics[width=0.8\linewidth]{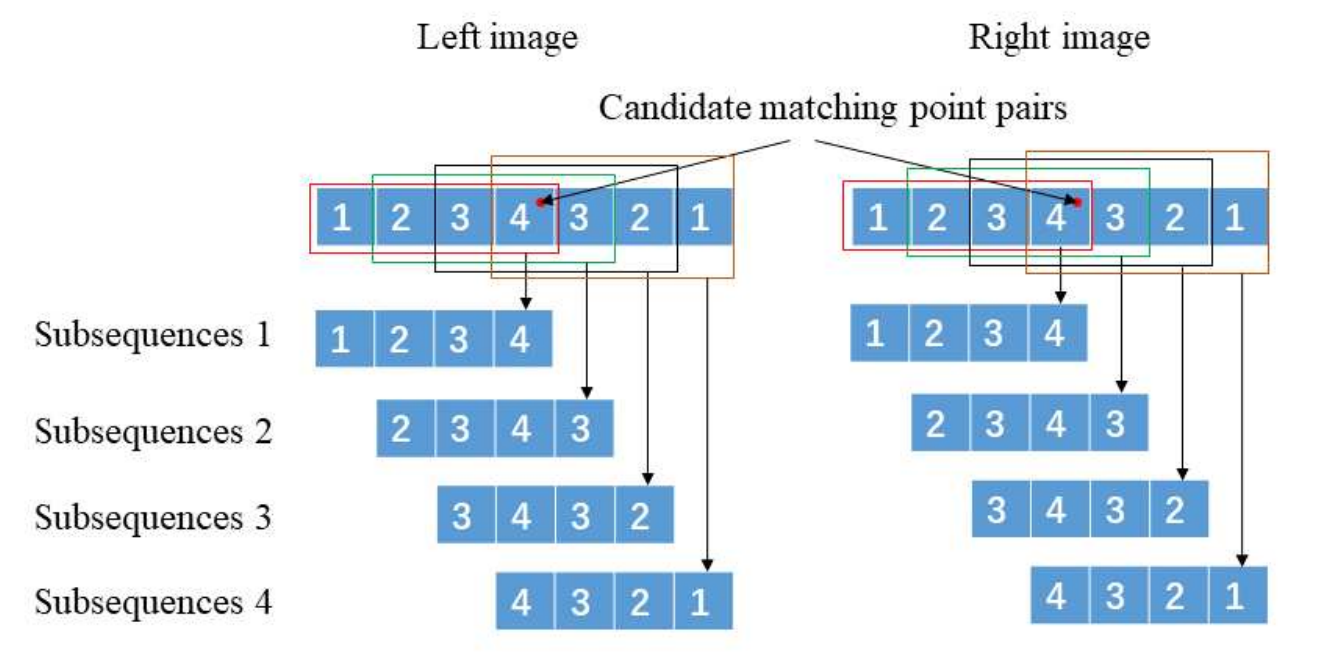}
\caption{DeBruijn analysis of refined matching verification.}
\label{duAnalysis}
\end{figure}

Each candidate matching pixel pair undergoes DeBruijn analysis. As shown in Fig.~\ref{duAnalysis}, a sliding window of length four is used to sequentially verify the decoded DeBruijn subsequences of the matching point pairs. If the subsequences differ, the candidate matching pair is discarded. 

To ensure that the cost values accurately reflect the correlation between pixels, the SGBM algorithm employs path integration of the matching cost in stereo vision across multiple directions. Let \( L_{r} \) represent a path traversed in the direction \( r \). The cost \( L_{r}(p, d) \) for pixel \( p \) at disparity \( d \) is recursively defined as:
\begin{equation}
   L_{r} \left(p , d\right) = C \left(p , d\right) + \min \left(L_{1}, L_{2}, L_{3}, L_{4}\right) - L_{5},
\end{equation}

\noindent where \( L_{1}, L_{2}, L_{3}, L_{4}, \) and \( L_{5} \) are the path costs corresponding to the neighboring pixel, defined as:
\begin{align}
    L_{1} &= L_{r} \left(p - r , d - 1\right) + P_{1}, \quad L_{2} = L_{r} \left(p - r , d\right),\nonumber\\
    L_{3} &= L_{r} \left(p - r , d + 1\right) + P_{1},\nonumber\\
    L_{4} &= \min \left(L_{r} \left(p - r , i\right)\right) + P_{2}, \quad i = d_{\min} \ldots d_{\max},\nonumber\\
    L_{5} &= \min \left(L_{r} \left(p - r , k\right)\right), \quad k = d_{\min} \ldots d_{\max}.\nonumber
\end{align}
Here, \( P_{1} \) and \( P_{2} \) are smoothness penalty coefficients. The term \( L_{5} \), representing the minimum path cost of the previous pixel, is introduced to prevent excessively large values in the calculations and ensure numerical stability. 

The aggregated matching cost is defined as:
\begin{equation}
S \left(p , d\right) = \sum_{r} L_{r}\left(p , d\right),
\end{equation}
\noindent where \( S(p, d) \) represents the total matching cost aggregated across all paths \( r \). The final disparity value \( D(p) \) is then determined by minimizing the aggregated cost:
\begin{equation}
    D \left(p\right) = \arg \underset{d}{\min}\textrm{ } S\left(p , d\right).
\end{equation}
\subsubsection{Point Cloud Generation}
\begin{figure}[tp]
	\centering
\includegraphics[width=0.7\linewidth]{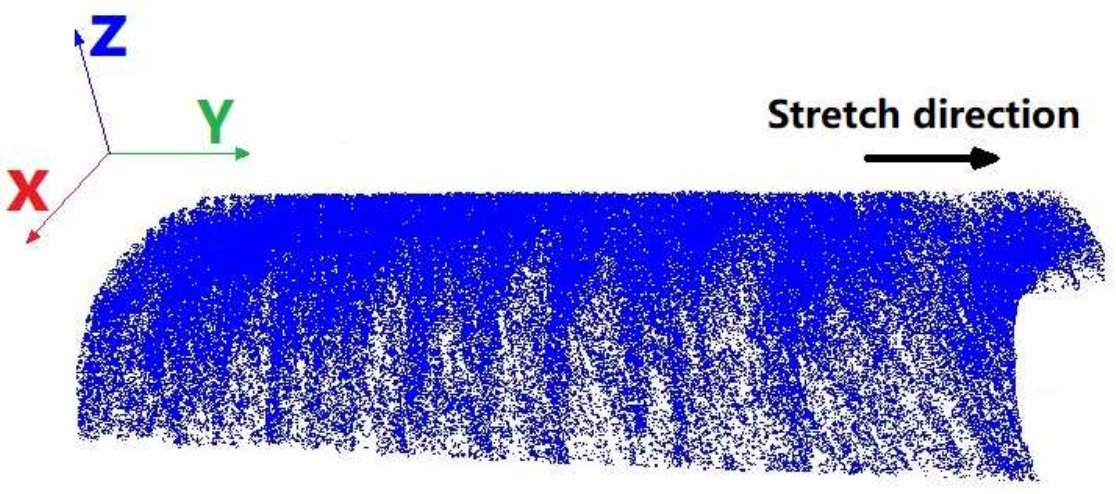}
\caption{The generated 3D point cloud of a deformable silicone object's surface.}
\label{pointcloud}
\end{figure}

Based on the disparity map, the 3D point cloud of the object's surface can be constructed, as illustrated in Fig.~\ref{pointcloud}.
Specifically, for each pixel \( p \) with coordinates \( (u, v) \) in the field of view, the 3D coordinates \((x_i, y_i, z_i)\) can be computed using the focal length \( f \), the optical center coordinates \( (c_x, c_y) \), the baseline distance \( B \), and the disparity \( D(p) \) from the left and right cameras as follows:
\begin{align}
    z_i &= \frac{f \cdot B}{D(p)},\nonumber\\
    x_i &= \frac{(u - c_x) \cdot z_i}{f},\\
    y_i &= \frac{(v - c_y) \cdot z_i}{f}.\nonumber
\end{align}

By iterating over all pixels in the disparity map \( D(u, v) \), the 3D spatial point cloud of the surface is generated. 
\subsection{Modified PointNet-Based Force Estimation}
\begin{figure}[htp]
	\centering 
    \includegraphics[width=1.0\linewidth]{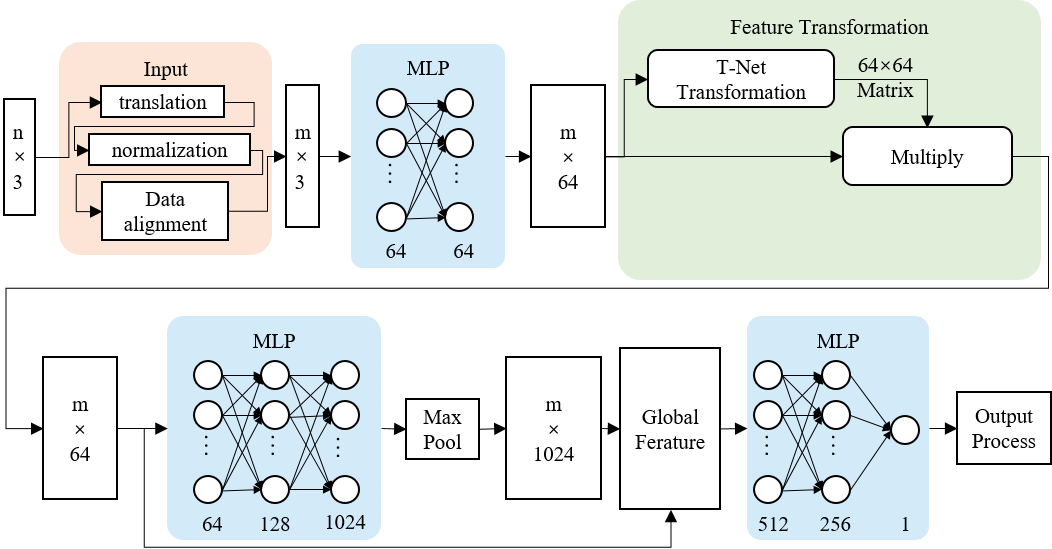}
	\caption{The modified PointNet network architecture.}
	\label{wangluo}
\end{figure}
The PointNet framework~\cite{qi2017pointnet} demonstrates exceptional capability in processing unstructured point cloud data. By independently applying multilayer perceptrons (MLPs) to individual points, it efficiently extracts local features, while leveraging global max pooling to capture global deformation characteristics. This design ensures invariance to the order of the input points. In particular, PointNet boasts an architecturally simple, computationally efficient, and highly flexible structure. However, PointNet was originally designed for classification tasks, rendering it inapplicable for force estimation, a fundamentally regression-oriented problem. Surgical force estimation, in particular, demands not only exceptional accuracy but also real-time performance to deliver high-quality haptic feedback to the operating surgeon. Additionally, the training and deployment of accurate force estimation models using the high-density point cloud data generated in Section III-A can be computationally intensive and time-consuming. 

To overcome these challenges, targeted modifications and optimizations are implemented in both the network architecture and the training algorithm, ensuring the framework aligns with the stringent requirements of surgical force estimation tasks.


\subsubsection{Input Preprocessing and Network Modifications} To facilitate consistent input representation for the network, a preprocessing step is introduced for the point-cloud data. This step involves normalizing each point cloud individually by translating its centroid to the origin of the coordinate system and scaling the point cloud to fit within a unit sphere.

The centroid coordinates \((x_c, y_c, z_c)\) of a point cloud are computed as follows: 
\begin{align}
x_c &= \frac{\sum_{i=1}^{n} m_i x_i}{\sum_{i=1}^{n} m_i},\nonumber \\
y_c &= \frac{\sum_{i=1}^{n} m_i y_i}{\sum_{i=1}^{n} m_i},\\
z_c &= \frac{\sum_{i=1}^{n} m_i z_i}{\sum_{i=1}^{n} m_i},\nonumber
\end{align}
\noindent where \(m_i\) represents the weight of the \(i\)-th point in the cloud, and \(n\) denotes the total number of points in the cloud. 

By translating and normalizing, the coordinates of the point cloud are transformed as follows:
\begin{align}
\overset{\sim}{x}_{i} &= \frac{x_{i} - x_{c}}{D_{\max}},\nonumber\\
\overset{\sim}{y}_{i} &= \frac{y_{i} - y_{c}}{D_{\max}},\\
\overset{\sim}{z}_{i} &= \frac{z_{i} - z_{c}}{D_{\max}},\nonumber
\end{align}
\noindent where \(D_{\max}\) represents the maximum bounding envelope distance of the point cloud, which is the largest Euclidean distance between any two points within the point cloud.

The architecture of the proposed force estimation network is shown in Fig.~\ref{wangluo}, comprising an input layer, convolutional layers, pooling layers, a feature aggregation layer, activation layers, MLPs, dropout layers, and an output layer. The output layer is modified to a fully connected network layer to accommodate the continuous regression problem.

The original PointNet network employs ReLU as the activation function, which outputs zero for negative inputs, potentially leading to inactive neurons (``dying ReLU"). This limits the network's ability to model complex nonlinearities, critical for capturing the mechanical properties of soft tissues.

To address this, the ELU is utilized, defined as:
\[
\textrm{ELU}(\epsilon) = 
\begin{cases} 
\epsilon, & \epsilon > 0 \\
\alpha (\exp(\epsilon) - 1), & \epsilon \leq 0
\end{cases}
\]
\noindent where \(\alpha > 0\) is a hyperparameter. ELU is continuous and differentiable, mitigating the vanishing gradient problem. For \(\epsilon>0\), it resembles ReLU, and for \(\epsilon\leq0\), it behaves like sigmoid/tanh, effectively combining their strengths. This adaptation improves the network's capacity to capture soft tissue mechanics, enhancing its suitability for force estimation tasks.

\subsubsection{Optimization and Adaptation of the Training Algorithm} The proposed model is trained using the Nadam optimizer, an enhancement of the Adam optimizer. While Adam's adaptive learning rate adjustment is effective in many scenarios, it may struggle with slow convergence or fail to precisely locate the optimal point. Nadam addresses these issues by integrating Nesterov momentum, which combines Adam’s adaptive learning rates with a lookahead mechanism to accelerate convergence and improve accuracy.

The original Adam updates for momentum and velocity are defined as:
\begin{align}
    m_{t} &= \beta_{1} m_{t-1} + (1 - \beta_{1}) g_{t}, \\
    v_{t} &= \beta_{2} v_{t-1} + (1 - \beta_{2}) g_{t}^{2}, \nonumber
\end{align}
\noindent where \(g_{t}\) is the gradient at time \(t\), \(\beta_{1}\) and \(\beta_{2}\) are the decay rates, and \(m_{t}\) and \(v_{t}\) represent the momentum and velocity updates, respectively.

In Nadam, Nesterov momentum introduces a refined momentum update, given by:
\[
\overset{\sim}{m}_{t} = \beta_{1} m_{t} + (1 - \beta_{1}) g_{t}.
\]

The bias-corrected estimates of momentum and velocity are computed as:
\begin{align}
    \hat{m}_{t} &= \frac{\overset{\sim}{m}_{t}}{1 - \beta_{1}^{t}},\\
    \hat{v}_{t} &= \frac{v_{t}}{1 - \beta_{2}^{t}}, \nonumber
\end{align}
\noindent where \(\beta_{1}^{t}\) and \(\beta_{2}^{t}\) represent \(\beta_{1}\) and \(\beta_{2}\) to the power of \(t\).

The parameter update rule for Nadam is:
\[
\theta_{t+1} = \theta_{t} - \eta \cdot \frac{1}{\sqrt{\hat{v}_{t}} + \varepsilon} \left( \beta_{1} \overset{\sim}{m}_{t} + \frac{(1 - \beta_{1}) g_{t}}{1 - \beta_{1}^{t}} \right),
\]
\noindent where \(\eta\) is the learning rate, and \(\varepsilon\) is a small constant to ensure numerical stability. By leveraging the benefits of both adaptive learning rates and Nesterov momentum, Nadam enhances optimization efficiency and achieves better performance for the force estimation network. For this regression task, the mean squared error (MSE) is employed as the loss function.

\section{Experimental Validation} 
In this section, to validate the effectiveness of the proposed VBFE scheme, traction tests were performed on a platform developed consisting of silicone intestinal models with three different stiffness levels and a commercial surgical robot.
\subsection{Experimental Setup}
\begin{figure*}[tp]
	\centering 
\includegraphics[width=0.65\linewidth,height=0.25\textheight]{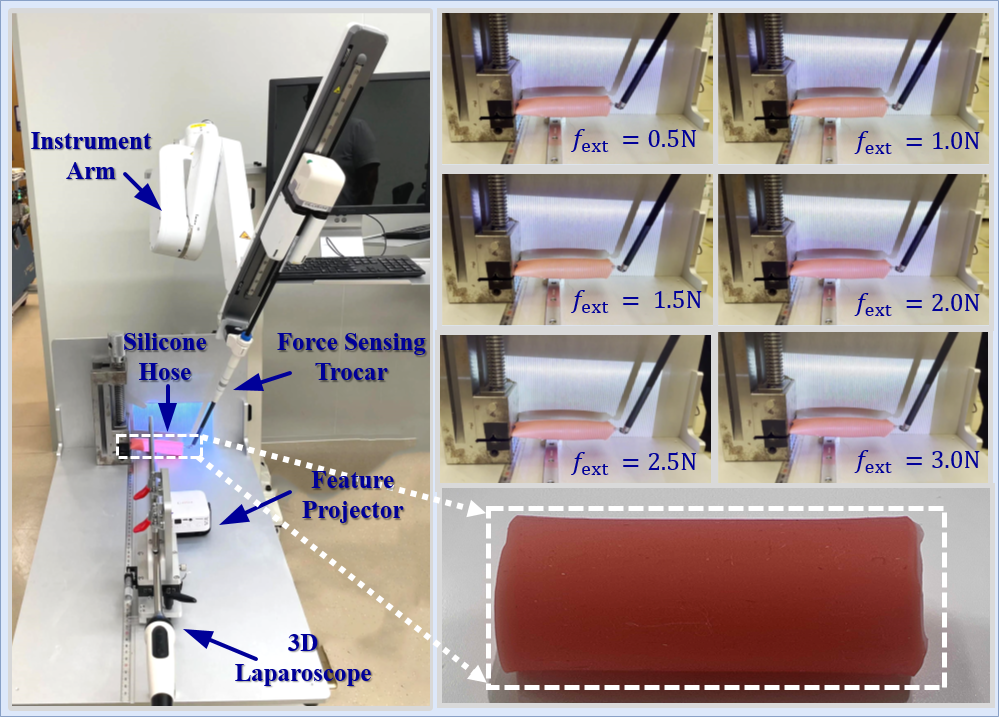}
\caption{The experimental setup. The experimental platform is shown on the left. The upper right illustrates snapshots of the pull-hold process under varying external forces (\(f_{\text{ext}}\)), ranging from 0.5 N to 3.0 N. The silicone hose, with calibrated stiffness, is displayed in the lower right.}
	\label{setup}
  \end{figure*}

  \begin{figure*}[!h]
	\centering 
\includegraphics[width=0.8\linewidth]{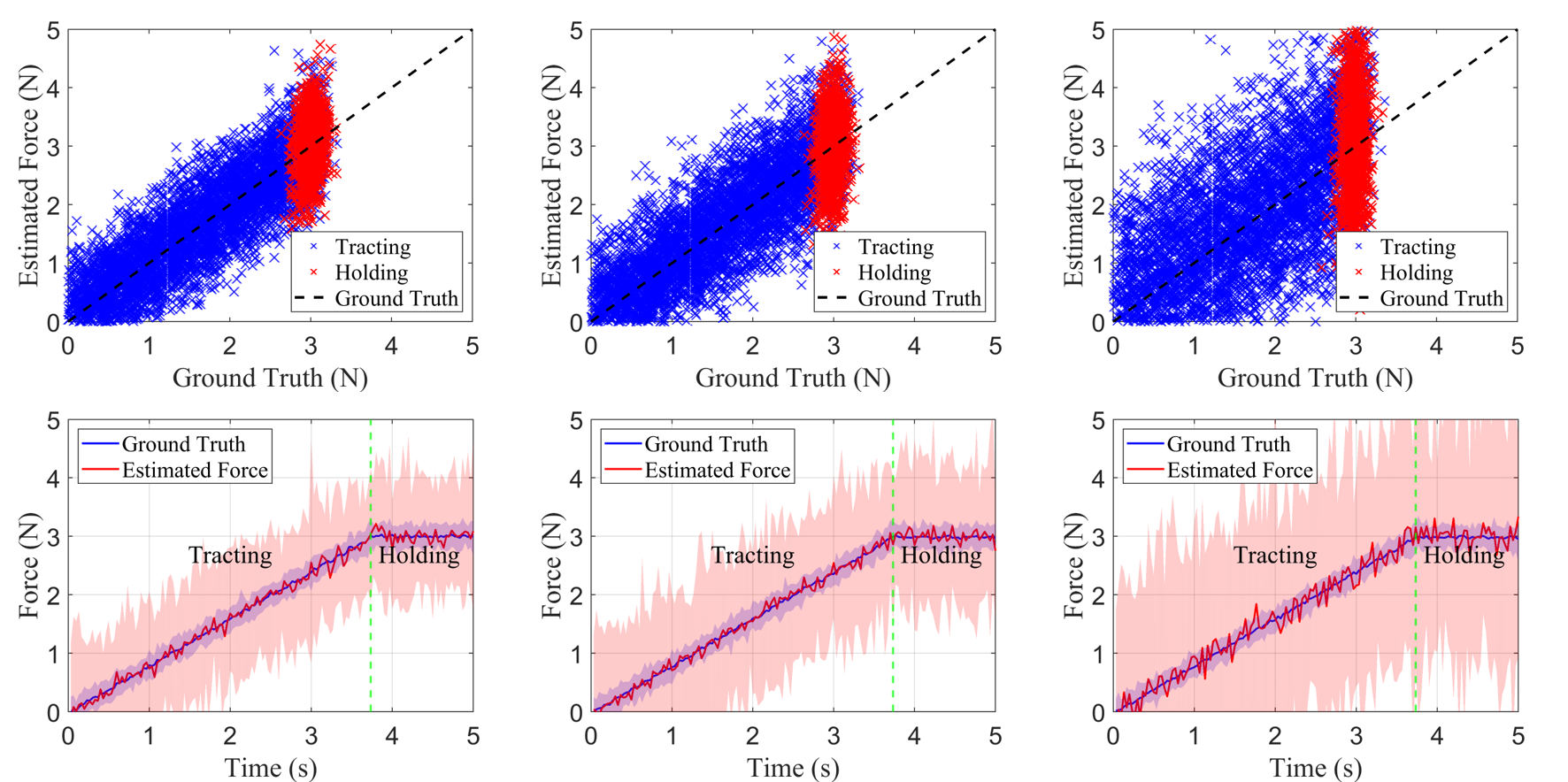}
\caption{The force estimation results. The subplots in the upper row present the scatter plots for estimated forces during the tracting (blue) and holding (red) phases against phase (blue) and holding phase (red), compared with the ground truth indicated by the black dashed line. The subplots in the lower row illustrate the mean force estimates (red) against the measurements (blue), with shaded regions indicating data intervals. From left to right, the subplots correspond to silicone samples with low, medium, and high stiffness levels, respectively.}\label{result2}
\end{figure*}

\begin{figure*}[!ht]
	\centering 
\includegraphics[width=0.8\linewidth]{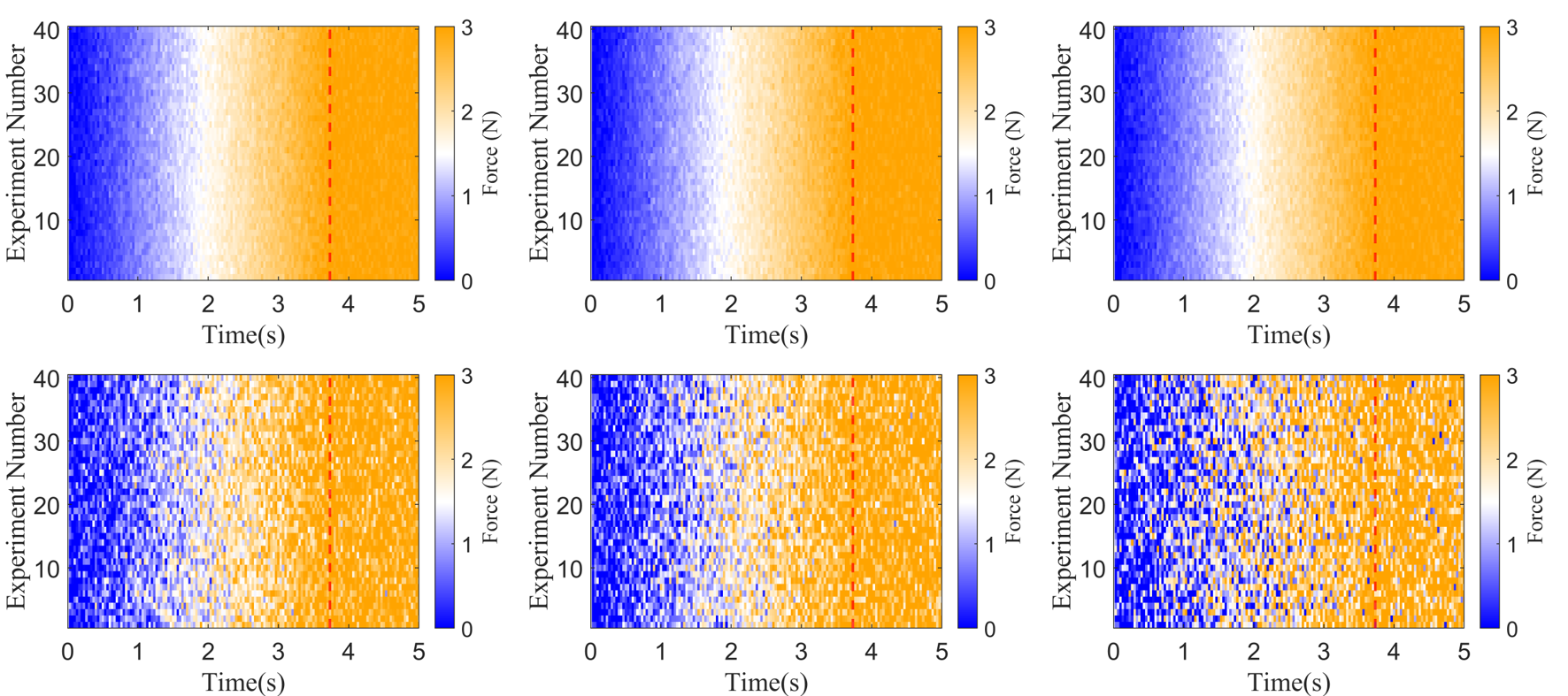}
\caption{The force estimation results. The top row shows the ground truth force measurements from the force-sensing trocar, while the bottom row illustrates the corresponding force estimates derived from visual deformation data. The three columns represent results for silicone objects with soft, medium, and hard stiffness levels, respectively. Each row shows one experiment. In each subplot, the colorur represents the force values (N) from low contact force (blue), medium contact force (white), and high contact force (orange). The red dotted lines differentiate between the tracting and holding phases.}\label{result}
  \end{figure*}

The experimental platform was constructed using the Toumai research kits provided by Shanghai Microport Medbot (Group) Co., Ltd.~\cite{wang2023autosurg}. As shown in Fig.~\ref{setup}, the platform includes a high-precision force sensing trocar (Model TRF85D) to measure the true interaction force and a surgical instrument arm (Model M0000339) equipped with a 3D electronic endoscope (Model EL824). The force-sensing trocar features a measurement range of ±5 N with a precision of 0.1 N. The robotic arm is equipped with grasping forceps (Model IN803A) that execute retraction and clamping operations in Cartesian space. These forceps are used to secure clamp one end of the silicone tube and perform the retraction operation. During the retraction process, the force sensor measures the interaction force applied to the silicone tube in real time, while the 3D electronic endoscope captures deformation images. These images are subsequently processed to generate a 3D point cloud for further analysis. The feature projector (Model L-mix), which implements structured light projection on the object surface. The entire experimental platform is designed to stably simulate the clinical operation environment, ensuring the reliability and repeatability of the experimental data. The three types of silicone materials, with stiffness values of 40 N/m, 80 N/m, and 200 N/m, represent soft, medium, and hard silicone tubes, respectively.

\subsubsection{Force Model Training}
A custom dataset has been collected using the Toumai commercial laparoscopic surgical robot affixed with a force-sensing trocar, along with the DFVision medical stereo endoscope and a silicone hose. This dataset has been open source on GitHub and is accessible at: https://github.com/CrisYaoMF/Force-Estimation-for-Soft-Tissue. As the manipulator stretches the silicone tube, the force-sensing trocar measures the applied force in real time, paired with the images of the deformation captured by the endoscope. The 3D laparoscope captures images of the silicone tube and generates 3D point cloud data at a sampling rate of 30 Hz. These paired data were fed into the proposed interaction force estimation network for training. The network was trained using the Keras framework, leveraging the Nadam optimizer to enhance both training speed and convergence. Throughout the process, the network weights were iteratively adjusted to minimize the RMSE between the force estimates and the actual measurements recorded by the force-sensing trocar, ensuring accurate model performance.

\subsubsection{Traction Experiment}
In the traction experiment, the manipulator gradually applies a force from 0 N to 3 N, holding the force steady once reached, with the entire process lasting 5 seconds. To ensure a uniform variation of force over time, different pulling speeds are employed for silicone tubes of varying stiffness: objects with stiffness levels of 40 N/m, 80 N/m, and 200 N/m are stretched at speeds of 0.01 m/s, 0.005 m/s, and 0.002 m/s, respectively. During the experiment, as the manipulator stretches the silicone tube, the 3D laparoscope continuously captures deformation images of the silicone tube and generates the corresponding 3D point cloud at a sampling rate of 30 Hz. Each experiment is repeated 40 times for each type of material to evaluate reliability and robustness.

\subsection{Experimental Results}
To evaluate the proposed force estimation scheme, the force estimation error statistics were quantified using the mean absolute error (MAE), mean squared error (MSE) and standard deviation (SD), denoted \(\sigma\), as shown in Table~\ref{tab.1}. The experimental results demonstrate a high level of accuracy in force prediction, particularly for lower stiffness conditions (40 N/m and 80 N/m), where prediction errors are relatively small, highlighting robust predictive precision.
\begin{table}[h]
\caption{Force estimation results with three stiffness}
	\centering
\label{tab.1}
\begin{tabular}{ccc}
\toprule  
\bf{Stiffness of the silicone tube}& \bf{MAE}$\left(\sigma\right)$  &  \bf{RMSE}$\left(\sigma\right)$ \\ \midrule
40N/m & 0.4055 (0.2558) & 0.3023 (0.3531) \\
80N/m & 0.4748 (0.3593) & 0.3544 (0.5047) \\
200N/m & 0.8044 (0.6108) & 1.0201 (1.4735) \\
\bottomrule
\end{tabular}
\end{table}
However, as the stiffness of the silicone material increases, the force estimation error also rises. For the object with a stiffness of 200 N/m, the MAE and MSE reach 0.8044 and 1.0201, respectively, with SD (\(\sigma\)) values of 0.6108 and 1.4735, significantly exceeding the errors observed for materials with stiffness levels of 40 N/m and 80 N/m. This increase in error can be attributed to the amplification of complex nonlinear characteristics in high-stiffness materials during deformation. Subtle errors in the deformation captured by visual data disproportionately propagate to interaction force estimates~\cite{penas2022unified}.

The force estimation results for the entire interaction operation are presented in Fig.~\ref{result2}. The figure demonstrates that the proposed method achieves highly accurate force estimation during both the traction and holding phases. The performance of the proposed force estimation method is consistent across these phases in terms of mean values and uncertainties, highlighting its robustness across different operational stages. From the perspective of linear correlation, as shown in the upper subplots, the force estimates maintain a strong correlation with the ground truth across materials with varying stiffness levels. However, high-stiffness materials tend to introduce larger uncertainties. This phenomenon is attributed to the smaller deformations exhibited by high-stiffness materials under the same applied force compared to soft-stiffness materials. These smaller deformations amplify the impact of any errors in the visual data, placing higher demands on the entire workflow of the vision-based force estimation method, including the quality of training data, the network's ability to model nonlinearity, and the resolution of the imaging system. Notably, in the context of MIS, the stiffness of human tissues typically ranges from 5 N/m to 50 N/m~\cite{singh2021mechanical}, which falls well within the effective operating range of the proposed method.

To validate the consistency of the force estimation algorithm, the force measurements and estimates from 40 repeated traction experiments are illustrated in Fig.~\ref{result}. The experimental results show that the proposed force estimation algorithm performs well for materials with soft (40 N/m) and medium (80 N/m) stiffness, where the estimated forces closely follow the trends of the ground truth across 40 repeated experiments, demonstrating good stability and repeatability. For materials with high stiffness (200 N/m), although the force estimates show slightly greater variability and reduced consistency, they still successfully capture the overall trend of force variation. 

\section{Conclusion}  \label{VII}
In this letter, a novel and effective binocular vision-based VBFE framework was developed for interaction with soft tissue in robotic-assisted surgeries. One-Shot structured light with a specially designed pattern and a two-step stereo vision technique was leveraged to achieve accurate and dense (pixel-wise) 3D point cloud of the surface of soft tissues to handle the reconstruction of smooth and texture-deficient surfaces. A modified PointNet-based force estimation method was developed by optimizing activation functions, loss functions, training algorithms, and the output layer, resulting in more adaptation to this complex nonlinear force model learning task. Traction experiments were conducted on a platform developed using a commercial surgical robot system and three silicone objects with different stiffness. The results evaluate the effectiveness and consistency of the proposed scheme, which highlights its potential application to force feedback in MIS. Compared to conventional methods, the hardware dependency of this scheme was reduced. For future work, the integration and miniaturization of the hardware platform require further research to enable the practical application of this method in commercial surgical robots.

\ifCLASSOPTIONcaptionsoff
  \newpage
\fi

\footnotesize
\bibliographystyle{IEEEtran}
\bibliography{ref.bib}

\end{document}